%% file: main.tex
\documentclass{ws-escri}
\usepackage{graphicx}
\usepackage[super,sort,compress]{cite}
\usepackage{lipsum}
\usepackage{xcolor}
\usepackage[T1]{fontenc}
\usepackage[inline]{enumitem}

\usepackage{glossaries}
\input{./glossaries.tex}

\begin{document}

\catchline{0}{0}{2019}{}{}

\markboth{Springer, Kyas}{Autonomous Drone Landing: Marked Landing Pads and Unexplored Lava Flows}

\title{Autonomous Drone Landing: Marked Landing Pads and Solidified Lava Flows}


\author{Joshua Springer}
\address{Department of Computer Science, Reykjavik University,\\
Menntavegur 1, 101 Reykjavik, Iceland\\
E-mail: joshua19@ru.is}

\author{Marcel Kyas}
\address{Department of Computer Science, Reykjavik University,\\
Menntavegur 1, 101 Reykjavik, Iceland\\
E-mail: marcel@ru.is}

%

\maketitle

\begin{abstract}
	\input{sections/abstract.tex}
\end{abstract}


\begin{multicols}{2}

	\section{Introduction}
		\input{./sections/introduction.tex}
	\section{Related Work}
		\input{./sections/related_work.tex}
	\section{Fiducial Landing}
		\input{sections/fiducial_landing.tex}
	\section{Lava Flow Landing}
		\input{sections/lava_flow_landing.tex}

	\section{Conclusions \& Future Work}
		\input{sections/conclusions_and_future_work.tex}

\bibliographystyle{plain}
\bibliography{references}

\end{multicols}
\end{document}

%% file: glossaries.tex
\newacronym{ENU}{ENU}{East, North, Up}
\newacronym{IMU}{IMU}{inertial measurement unit}
\newacronym{GPS}{GPS}{global positioning system}
\newacronym{GUI}{GUI}{graphical user interface}
\newacronym{IR}{IR}{infrared}
\newacronym{IRC}{IRC}{International Conference on Robotic Computing}
\newacronym{LIDAR}{LiDAR}{light detection and ranging}
\newacronym{RAVEN}{RAVEN}{Rover-Aerial Vehicle Exploration Network}
\newacronym{RGBD}{RGBD}{RGB + depth}
\newacronym{ROS}{ROS}{Robot Operating System}
\newacronym{RTK}{RTK}{real time kinematic}
\newacronym{DoF}{DoF}{degrees of freedom}
\newacronym{CNN}{CNN}{convolutional neural network}
\newacronym{DTM}{DTM}{digital terrain model}

%% file: sections/abstract.tex
Landing is the most challenging and risky aspect of multirotor drone flight,
and only simple landing methods exist for autonomous drones.
We explore methods for autonomous drone landing in two scenarios.
In the first scenario, we examine methods for landing on known landing pads using
fiducial markers and a gimbal-mounted monocular camera.
This method has potential in drone applications where a drone must land more accurately
than GPS can provide (e.g.~package delivery in an urban canyon).
We expand on previous methods by actuating the drone's camera to track the marker over time,
and we address the complexities of pose estimation caused by fiducial marker orientation ambiguity.
In the second scenario, and in collaboration with the RAVEN project,
we explore methods for landing on solidified lava flows in Iceland,
which serves as an analog environment for Mars
and provides insight into the effectiveness of drone-rover exploration teams.
Our drone uses a depth camera to visualize the terrain,
and we are developing methods to analyze the terrain data for viable landing sites in real time
with minimal sensors and external infrastructure requirements,
so that the solution does not heavily influence
the drone's behavior, mission structure, or operational environments.


%% file: sections/introduction.tex
\label{section:introduction}

Drones are popular solutions for automating aerial tasks.
For example, package delivery by multirotor drones has the potential to revolutionize urban shipping
by providing a cheap way to decrease delivery times
and reduce dependence on roadway infrastructure.
Drones have also impacted the sector of remote sensing,
functioning as a ``tripod in the sky'' to which users can attach their preferred sensors,
thereby advancing infrastructure inspection, surveillance, search and rescue, archaeology, geology,
and even Mars exploration.
The main benefits of drones are
\begin{enumerate*}
\item the removal of human operators
from the task loop, which increases safety and reduces costs, and
\item the ability to quickly get to hard-to-reach places.
\end{enumerate*}

Removing the human operator means that the drone must function autonomously by
generating and executing its mission
or semi-autonomously by executing a mission already generated by a human.
We can break down such missions into a few elementary steps:
takeoff,
waypoint-to-waypoint flight,
triggering of external tasks,
and landing.
The autonomous execution of many of these components is already reliable.
Autonomous takeoff is done with an \gls{IMU} and either \gls{GPS} or optical flow
for position maintenance once in the air.
The drone can fly from waypoint to waypoint using \gls{GPS},
dead-reckoning via optical flow, or identification of landmarks in the environment.
Time-based or location-based triggers activate secondary tasks,
e.g.~collection of pictures or videos,
payload jettison/collection,
signaling via radio or light, etc.,
whose execution can be handled by a companion board outside of the real-time environment
of the flight controller.

Landing, however, still needs sufficient methods, particularly in harsh, antagonistic environments.
Typically, drones carry out autonomous landings by navigating
to a particular \gls{GPS} coordinate representing a known-safe location
and descending in place until reaching the ground
without sensing obstacles.
More sophisticated methods integrate a rangefinder to estimate the altitude of the drone
relative to the local terrain instead of sealevel (as with \gls{GPS}).
Others add horizontal object avoidance (possibly only from front-facing sensors)
or use \gls{IR} beacons to guide the drone to a landing site with active infrastructure.
Many peers contributed to this field (see Section~\ref{section:related_work}),
using combinations of \gls{GPS}, radio beacons, \gls{IR} beacons,
visual markers and positioning systems in the case of landing on a known, structured
landing site.
Others contributed to evaluating new,
unstructured landing sites
in unexplored terrain, using \gls{LIDAR}, \gls{RGBD} cameras,
or structure-from-motion to create a 3D view of potential landing sites for analysis.
These projects are convincing proofs of concept for sophisticated landing methods
but are limited.

We present our approaches
to improve autonomous drone landing methods in structured
and unstructured scenarios.
We minimize the required sensors, ground infrastructure, behaviors, and power consumption
the drone needs to identify a landing site and autonomously land safely.
Our approach allows the methods to generalize to many different platforms and mission types
where e.g.~certain infrastructure is unavailable, has prohibitive latency,
or has stringent power requirements.

In the structured case, we visually mark the landing sites with \emph{fiducial markers} (Section~\ref{section:fiducial_landing}).
We improve on previous methods by actuating the drone's camera to track the marker
instead of using the fixed, downward-facing camera,
but this requires pose transforms that consider the marker's orientation, which is ambiguous
and leads to erroneous control signals.
We explore options for filtering the marker pose
and deriving the landing pad's orientation using the camera.
Such methods are useful with known landing sites, 
with or without \gls{GPS}.

In the unstructured case,
we take our exciting application scenario from our partners in the \gls{RAVEN}~\cite{raven_whitepaper}
project,
who are testing collaborative drone-rover teams for Mars exploration,
using Iceland as an analog environment (Section~\ref{section:lava_landing}).
The task of the drone in this context is to collect data and samples in areas a Mars rover
cannot reach, i.e.~on top of a solidified lava flow.
Thus, the drone must autonomously land in a harsh environment
with no external infrastructure for positioning or sophisticated computing.
Our ongoing task is to develop a method for terrain analysis
to enable the drone to identify viable landing sites in lava flows.
We use state-of-the-art computing hardware to analyze the terrain
onboard our small drone in real-time, whereas previous methods required large drones or could not run in real-time.
We use an \gls{RGBD} camera to provide a 3D view of the terrain below the drone
and detect fiducial markers.
We collect data and will test our methods in two Icelandic lava fields
-- Holuhraun and those near the town of Hafnarfjör{\dh}ur, providing insight into
the challenges of autonomously landing in a terrestrial or planetary lava field.

We measure our success by a mission where an autonomous drone takes off
from a fiducially-marked landing pad,
flies over a nearby lava field,
identifies a safe landing site,
and executes an autonomous landing.
Finally, it should take off again and land at the original takeoff site, without a human operator.
This will prove both methods in real-world landing scenarios with a small, unified sensor set.

This paper is an extension of a PhD workshop paper
published in the 2022 \gls{IRC}\cite{irc_phd_workshop_paper},
with more details on the related work,
our drone platform,
new terrain data
corresponding processing methods,
and our the anticipated real-world testing scenarios.

%% file: sections/related_work.tex
\label{section:related_work}

Many peers have explored various methods for autonomous landing,
leveraging combinations of \gls{GPS}, ground-based infrastructure, and onboard sensors
to identify a landing site.
Here, we examine methods that use \gls{GPS} only as a secondary navigational tool or not at all.
Even with the low positional error that comes with \gls{RTK} systems, \gls{GPS}-based methods are blind
to the environment and only applicable in a terrestrial setting.
We prefer to study methods that evaluate the landing site to some extent,
such that obstructions to the landing site will stop the autonomous landing process
instead of potentially resulting in a crash.

\subsection{Fiducial Methods}

Several projects have accomplished fiducial landing by marking a landing pad with fiducial markers
and locating it with one or more RGB cameras onboard the drone.
In most cases, the cameras are fixed rigidly
or mounted on a gimbal with a fixed -- but stabilized -- orientation.
Fixed cameras simplify the pose estimation,
allowing the drone to use the camera's known orientation in its pose estimation
rather than the unreliable orientation of the marker (see Section~\ref{section:fiducial_landing}).

Araar et~al.\ use two cameras that are rigidly fixed to a drone
to land on a moving landing pad with multiple April Tag markers of different
sizes~\cite{fiducial_landing_many_markers_voting_fixed_camera}.
This increases detection range and robustness and allows a voting scheme to help overcome
erroneous pose estimates.
Chaves et~al.\ describe a method that uses a single, fixed, front- and downward-facing camera
to identify April Tag markers and land on a ship at
sea~\cite{fiducial_landing_ship_6dof_single_fixed_downfront_camera_apriltag}.
Liao et~al.\ use April Tags to designate stationary
landing pads, and multiple drones identify them with fixed cameras while cooperating to land
with a priority-based order~\cite{fiducial_landing_two_fixed_cameras_apriltag}.
Nguyen et~al.\ use a single camera
mounted on a gimbal, pointing $90^\circ$ downwards during the landing to identify their
high-visibility marker for nighttime landing on a stationary target~\cite{fiducial_landing_downward_facing_90_deg_gimbaled_camera}.
Polvara et~al.\ describe a method for landing a drone with two fixed cameras
on the deck of a pitching and rolling ship
with a landing pad marked with an 
AR Tag\cite{ar_tag,fiducial_vessel_landing_ar_tag_two_fixed_cameras}.
Many such methods report visual loss of the marker due to the drone's changing orientation
during the landing, leading to restarts of the landing.

Kim et~al.\ make the landing pad identification more robust
by using an omnidirectional camera pointed down~\cite{omnidirectional_fiducial_landing},
and accomplish autonomous landing with their own simple, large, red marker.
However, the extreme peripheral distortion of the fisheye lens
prevents identifying the landing pad with more complex visual markers if the landing pad
is not near the center of the camera frame.
Another solution, presented by Tanaka et~al., uses an actuated, gimbal-mounted camera to
track their \emph{proprietary} marker system during landing~\cite{lentimark_landing}.
Their marker system extends the AR Toolkit with external visual patterns to reduce the effect of
orientation ambiguity, so that the marker's 6-\gls{DoF} pose can be determined accurately
-- a problem present in typical fiducial systems.

We improve on these results by achieving autonomous fiducial landing with \emph{open-source}
software only and using only a single, actuated camera.
This requires addressing the problem of fiducial marker orientation ambiguity in the case of
open-source fiducial systems, either through filtering or by using the known orientation
of the gimbal-mounted camera.

\subsection{Terrain Analysis Methods}

Analyzing the terrain under the drone allows the drone to land in
unknown, unprepared sites.
Several relevant sensors (e.g.~\gls{LIDAR}, \gls{RGBD} cameras, optical flow sensors,
and conventional monocular and stereo cameras)
allow the drone to distinguish regions that are rigid/dynamic, smooth/rough, or level/slanted.
There are intrinsic difficulties in optically sensing self-similar terrain
in running real-time algorithms
(e.g.~analysis of dense terrain reconstruction from \gls{LIDAR} or \gls{RGBD} sensors).

Desaraju et~al.~\cite{rooftop_landing} present a method for successful rooftop landing;
the drone's onboard Odroid U2 computer locates landing sites onboard at a rate of 1 Hz.
Pluckter and Scherer present a method where a drone
takes a series of pictures throughout takeoff,
conducts a mission,
then follows its mission trajectory backward and attempts to
land at its takeoff location by matching the terrain to the pictures it collected during ascent~\cite{drone_landing_unstructured_environments}.
Garg et~al.\ present a method of both monocular
and stereo visual terrain analysis that can densely reconstruct the scene but is also able
to detect non-rigid surfaces (e.g.~water) using optical flow~\cite{monocular_stereo_cues_landing_site_evaluation}.
Putranto et~al.\ present a method for finding safe landing sites
with U-net and RGB input images\cite{slz_identification_semantic_contour}.
While the method identifies safe landing sites, it has a slow framerate even on non-embedded hardware,
showcasing the difficulty of running such methods in real-time onboard a drone.
Maturana and Scherer's method~\cite{conv_3d_lidar_landing} quickly finds landing sites
in \gls{LIDAR} point clouds using a \gls{CNN}
trained to high accuracy on synthetic data sets taken from real-world scenes.
It distinguishes between compliant objects and ground points
to determine the landing safety of each area.

We address a new autonomous landing scenario in the static
but harsh environment of solidified lava flows,
with an \gls{RGBD} camera for terrain detection
and focus on real-time, embedded processing.

%% file: sections/fiducial_landing.tex
\label{section:fiducial_landing}

Fiducial markers provide a convenient way to visually identify a marked landing pad
using a drone's monocular camera.
First, we created a method for fiducial landing
that actuated the drone's camera to track the marker independently of the drone's orientation.
This makes it easier for the drone to estimate the relative pose (position and orientation)
from itself to the landing pad. In contrast, previous work often reported a visual loss of the marker
due to the camera being rigid on the drone or pointing vertically down.
The position component of a marker's pose is the distance vector from the camera's center
to the center of the marker in the camera's coordinate system.
This is a reliable datum with some noise in many fiducial systems.
However, since the camera is actuated, we must consider the marker's position with
either its orientation or the camera's orientation to calculate where the landing pad is.
Following our preference for minimal sensor requirements, we preferred to use the marker's orientation,
with the warranted assumption that its actual orientation is flat on the ground
and facing up.
(To use the camera's orientation requires an additional \gls{IMU}.)

Initial, simulated tests of fiducial landing with an actuated camera showed that
the orientation ambiguity of fiducial markers resulted in erroneous control signals
when the marker orientation was used in pose transforms~\cite{joshua_master_thesis}.
This motivated our evaluation of several fiducial systems
in terms of their prevalence of orientation ambiguity~\cite{fiducial_precursor_evaluation}
(see Section \ref{section:evaluation_of_orientation_ambiguity}).
We also created three derivative fiducial systems to mitigate the orientation ambiguity
and tested the systems in terms of orientation ambiguity and detection rate.
Real-world landing tests (see Section \ref{section:real_world_landing_tests}) conducted with the evaluated markers revealed an inverse relationship
between the prevalence of each system's orientation ambiguity and its landing accuracy,
defined in this case as the distance from the camera to the center of the marker after touchdown.
They also showed that, although the actuated camera setup can introduce erroneous control signals,
autonomous landing is still possible.
Moreover, searching for the landing pad can be as simple as spinning the drone in place
and tilting the camera up and down
in a \gls{GPS}-denied environment.

\subsection{Fiducial System Evaluation}
\label{section:evaluation_of_orientation_ambiguity}

Previously, we describe an evaluation of five fiducial systems in terms of orientation ambiguity
and detection rate
to determine their suitability for fiducial landing \cite{fiducial_precursor_evaluation}.
We tested the systems with \gls{ROS} on a Raspberry Pi 4 with 2 GB of RAM.
The systems process a 480p video stream from a
``Creative Technology Live! Cam Sync 1080p''
with a 77$^\circ$ field of view.
In order to quantify each system's orientation ambiguity,
we choose a corresponding marker
and move the camera around it using different slow and controlled motions:
translation, rotation, orbiting both up/down and left/right.
We detect spikes in each system's sensed linear and angular velocities,
which correspond to erroneous determinations of the marker's orientation.
All five fiducial systems process the same video.
Moreover, all markers appear in the video during all frames.
To quantify the detection rate of each system,
we film one marker
at distances of 1--3 meters and deflections of $0^\circ$ and $30^\circ$
to avoid interference as each system looks for white and black regions (Figs.~\ref{figure:apriltag} and~\ref{figure:whycode}).
The detection rate is the number of detections divided by the length of the video.

\vspace{0.1cm}
\begin{figurehere}
	\centering
	\includegraphics[width=0.325\linewidth]{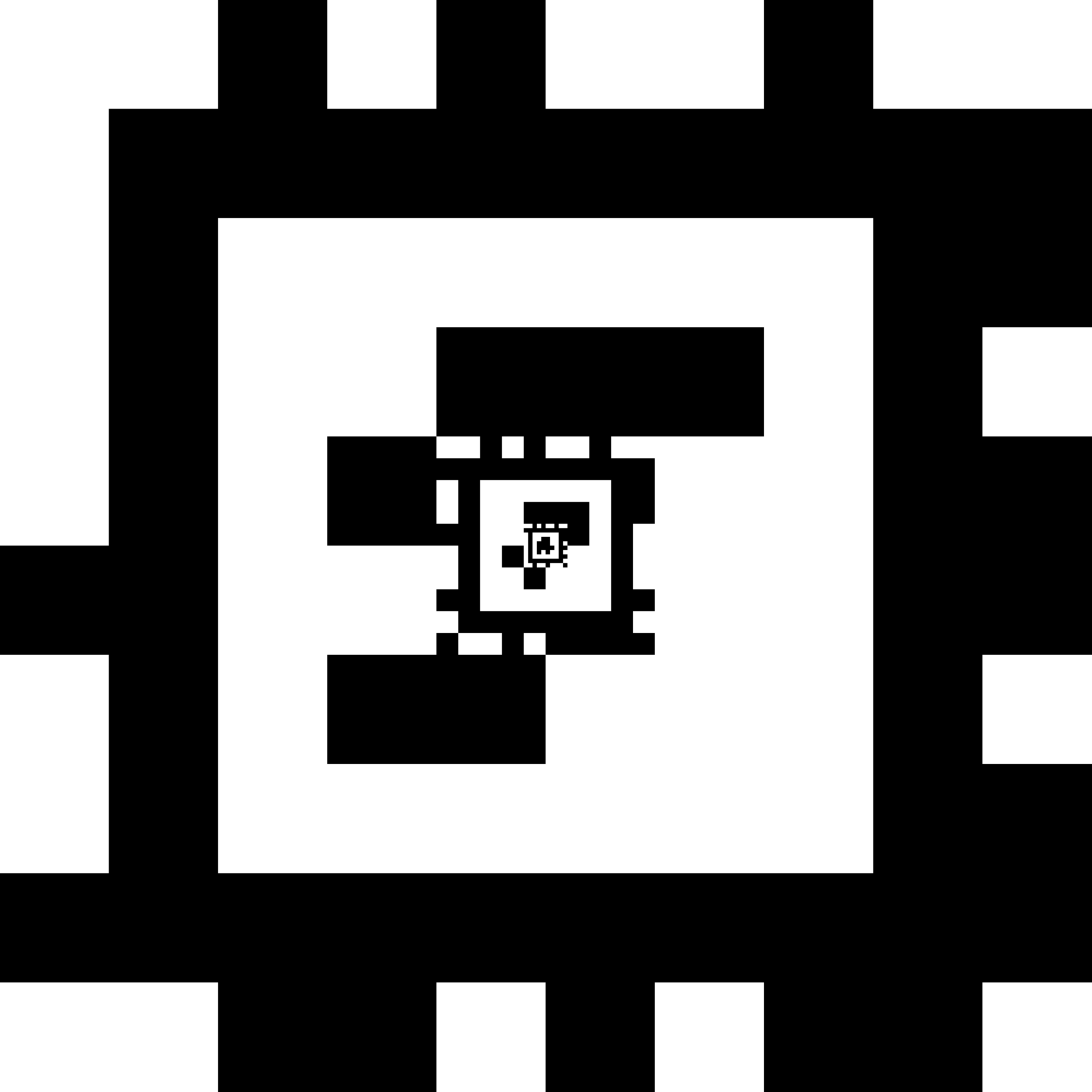}\quad
	\includegraphics[width=0.325\linewidth]{./images/tagCustom24h10_00002_00001_00000.pdf}
	\caption{April Tag 48h12 (left) and April Tag 24h10 (right) marker arrangements with embedded IDs 2, 1, and 0 in order of decreasing size.}
	\label{figure:apriltag}
\end{figurehere}
\vspace{-0.35cm}
\begin{figurehere}
	\centering
	\includegraphics[width=0.325\linewidth]{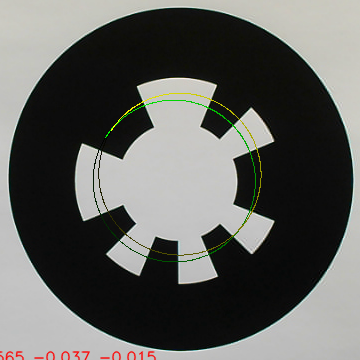}\quad
	\includegraphics[width=0.325\linewidth]{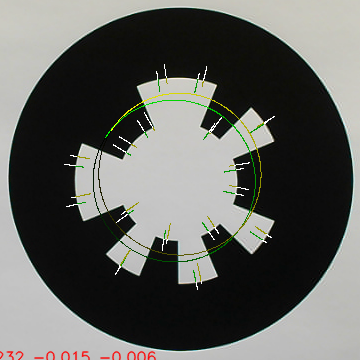}
	\caption{WhyCode marker with 8-bit ID 20. Identification with original WhyCode method (left),
	and with WhyCode Ellipse method (right).}
	\label{figure:whycode}
\end{figurehere}
\vspace{-0.35cm}

The first system is April Tag 48h12 (Fig.~\ref{figure:apriltag})~\cite{apriltag3_paper}.
It has 48 black and white ID bits, a minimum Hamming distance of 12 between all marker ID pairs,
for a total of 42,211 markers in the marker family.
It has an undefined center where it is possible to embed other,
smaller markers with different IDs, so that the drone maintains a pose estimate
for the landing pad, even as it is too close to identify the outer marker.

The second system is April Tag 24h10 (Fig.~\ref{figure:apriltag}),
which we implemented with the April Tag Generation library~\cite{apriltag3_paper}.
It is similar to April Tag 48h12, but has fewer ID bits, only 18 markers in the family
(adequate for our purposes), and requires less RAM.
We test this marker family to see if it can run faster on embedded hardware.

The third system is WhyCode~\cite{whycode_paper} (Fig.~\ref{figure:whycode}),
a lightweight fiducial system based on concentric black and white disks with a Manchester-encoded
binary ID.
The system determines two possible solutions for the marker orientation using the semi-axes of the
marker's ellipses under projection.
Then samples for the ID along the corresponding predicted
ellipses, here shown by the yellow and green ellipses over the marker.
It chooses the better aligned solution by minimizing the arclength intersection of the sampling
ellipse with each tooth.

The fourth system is our WhyCode Ellipse (Fig.~\ref{figure:whycode}),
and it adds a second phase of sampling to the otherwise unchanged WhyCode method.
The second sampling phase predicts the white-to-black transitions going radially outward from the center
of the marker on each ID bit,
and it chooses the solution that is more centered by this metric, in an effort to give
a more reliable decision.

We implemented a fifth system, called WhyCode Multi,
which calculates the pose of a set of 3+ WhyCode markers,
using the average position of the markers as the position of the bundle,
and using the plane connecting the markers as its orientation.

All of the systems exhibit orientation ambiguity, which manifests as sign flips in
the marker's orientation.
These propagate through pose transforms that calculate the position of the camera within the marker,
thus preventing the systems from reliably estimating the position of the landing pad at every step.
Fig.~\ref{figure:discontinuities_evaluation} shows an example of the effects of marker orientation
ambiguity.
For a single test case, it contains the \emph{position targets} in a relative \gls{ENU} coordinate
system that is centered on the marker, such that the \emph{East}, \emph{North}, and \emph{Up}
components represent the distance that the drone has to move left/right, forward/backward, and up/down
respectively.
Orientation ambiguity causes these position targets to have sign flips, denoted by dotted lines
and corresponding to spikes -- i.e.~\emph{discontinuities} -- in the east and north position targets.

\vspace{0.1cm}
\begin{figurehere}
	\centering
	\includegraphics[width=\linewidth]{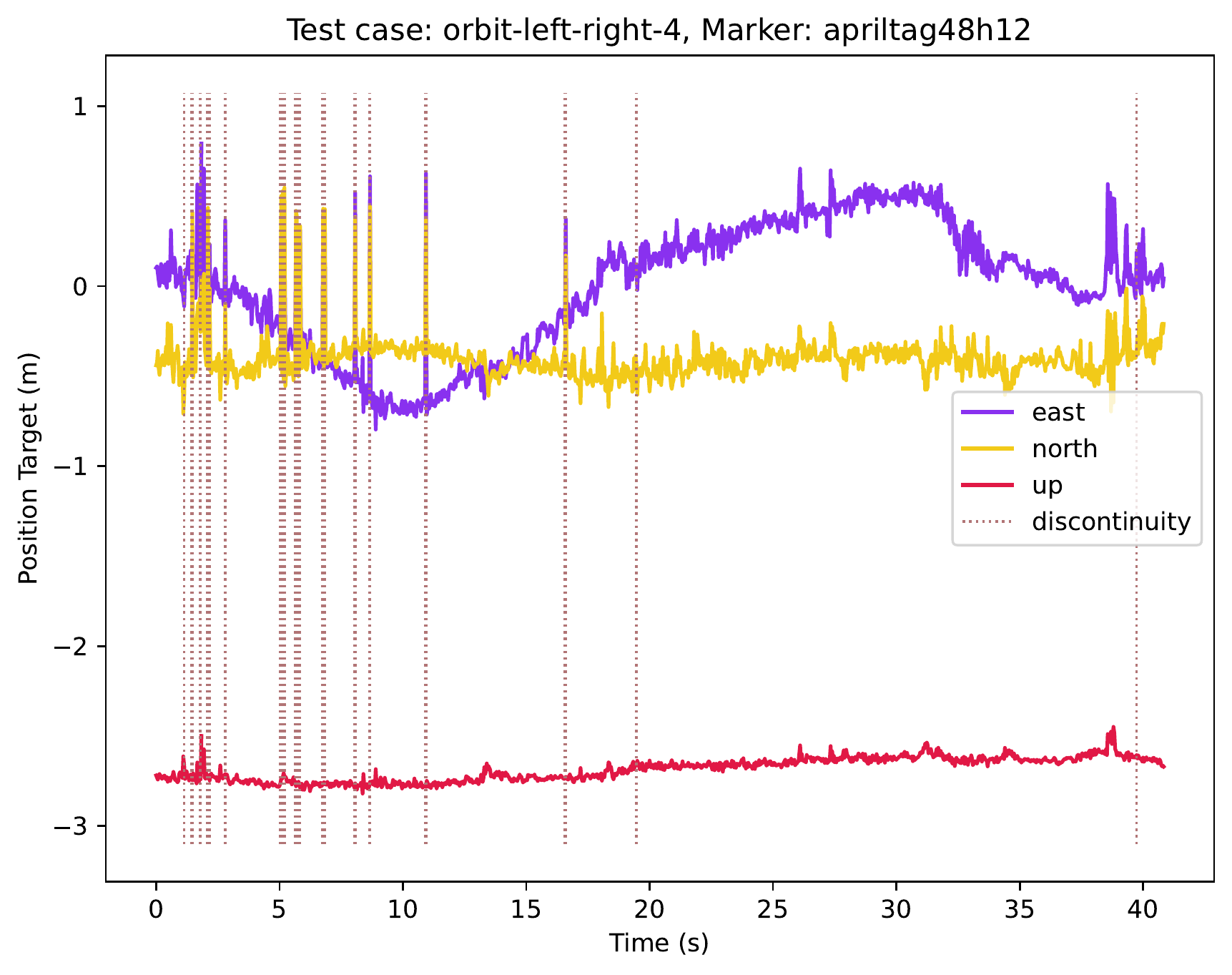}
	\caption{Example of orientation ambiguity causing discontinuities in the position estimate
	of the camera in the marker coordinate system.}
	\label{figure:discontinuities_evaluation}
\end{figurehere}
\vspace{-0.25cm}

The discontinuity rates for each system are shown in Fig.~\ref{figure:discontinuity_rates},
showing that the extra sampling phase in the WhyCode Ellipse system
decreases its discontinuity rate over the original,
that the WhyCode Multi system and April Tag 24h10 systems do not offer improvements.
The results also suggest that WhyCode Ellipse and April Tag 48h12
should be the best candidates for real world testing.
Figure~\ref{figure:detection_rates} shows the detection rates for each system.
The lightweight WhyCode systems run the fastest, with a small slowdown as a result of the
extra sampling phase in WhyCode Ellipse.
The WhyCode Multi system has a bimodal distribution, likely owing to sporadic identification of 
a single marker in the bundle, such that only two markers were sometimes identified, and therefore
no unique plane could be regressed between them.
The April Tag systems show the slowest detection rates, as a result of their longer \gls{ROS} pipeline
which includes an \texttt{image\_rect} node to remove camera distortion from the image,
whereas the WhyCode system does this natively.
All of the detection rates are acceptable for generating position target commands at 10 Hz.

\vspace{0.1cm}
\begin{figurehere}
	\centering
	\includegraphics[width=0.9\linewidth]{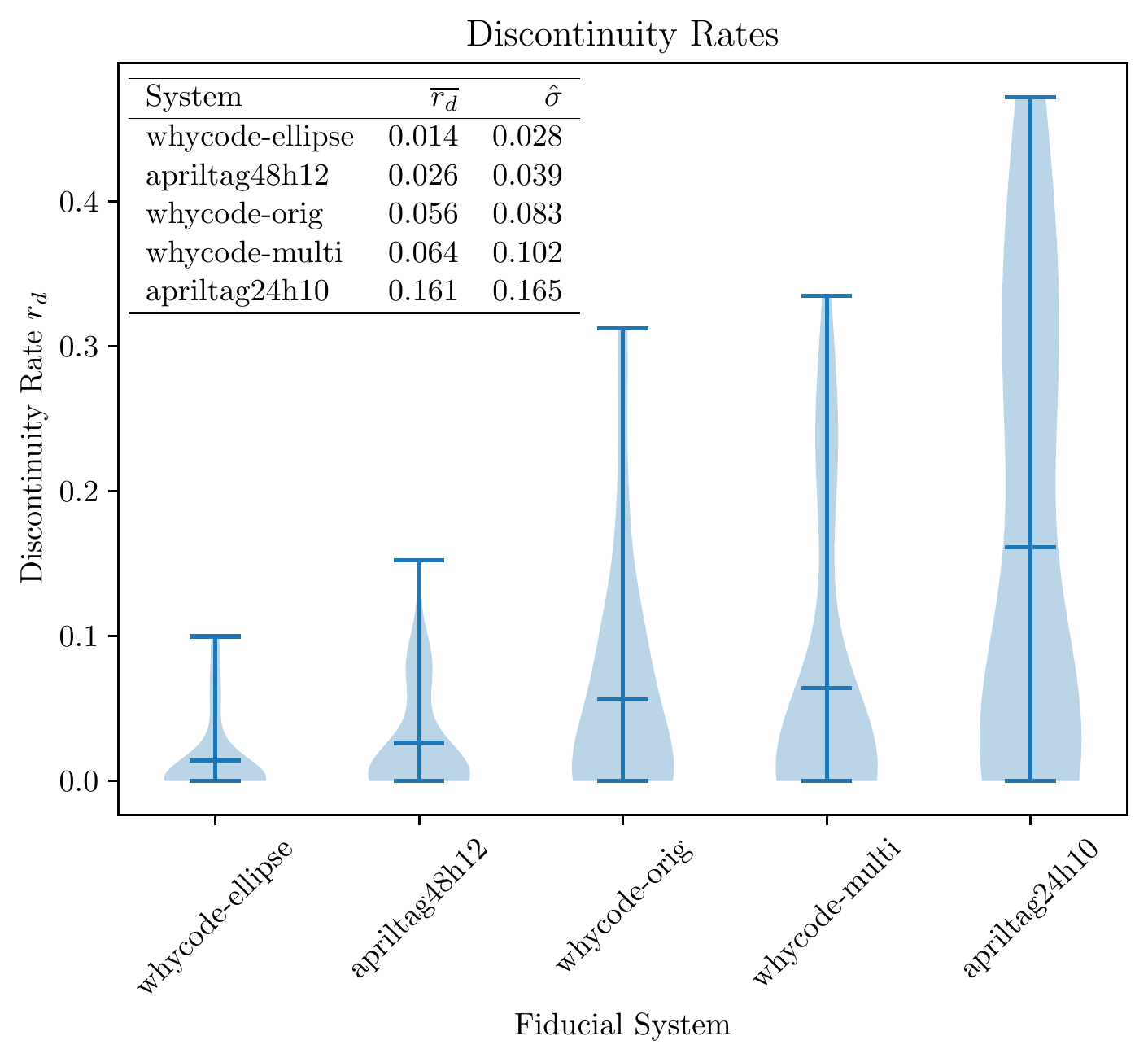}
	\caption{Distributions of discontinuity rates.}
	\label{figure:discontinuity_rates}
\end{figurehere}
\vspace{-0.35cm}
\begin{figurehere}
	\centering
	\includegraphics[width=0.9\linewidth]{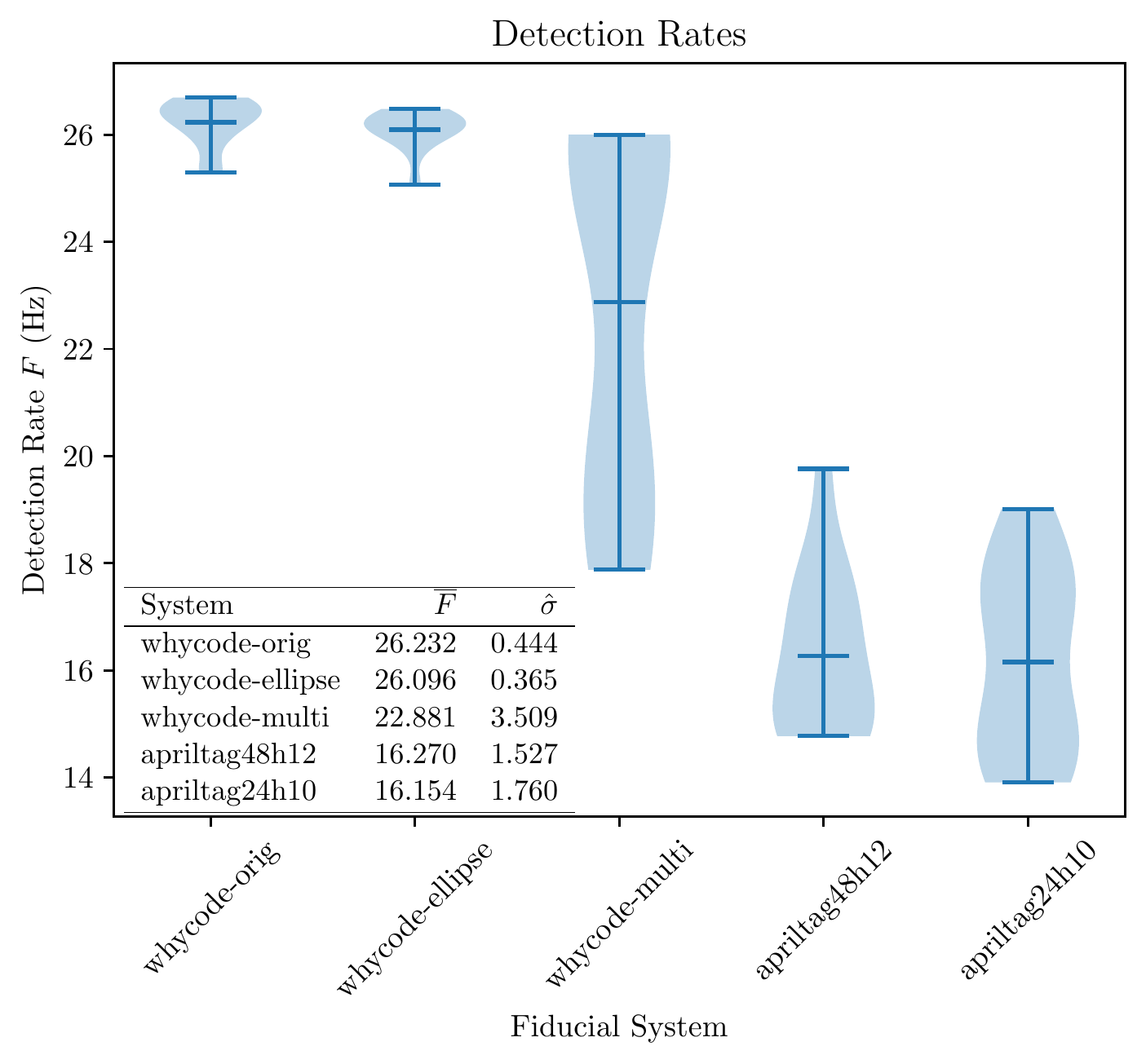}
	\caption{Distributions of detection rates.}
	\label{figure:detection_rates}
\end{figurehere}
\vspace{-0.35cm}

\subsection{Real World Landing Tests}
\label{section:real_world_landing_tests}

In previous work~\cite{fiducial_landing_paper},
we describe the real world testing of our fiducial landing algorithm in a \gls{GPS}-denied
environment, using the fiducial systems
described in Section~\ref{section:evaluation_of_orientation_ambiguity}.
We chose the small DJI Spark as a testing platform because it is safe to fly inside
wthout damaging itself or other things,
and we anticipated that it would occasionally receive erroneous control signals.
It also has sensors to estimate its speed and avoid obstacles,
and it gives access to the DJI Mobile SDK for automation,
such the experiments can generalize to other DJI drones with minimal changes.
The DJI Mobile SDK uses an app-style architecture, such that a tablet or phone
can provide a \gls{GUI} and access to the drone's video feed, in-flight data, and send control commands.
However, the DJI Spark has no onboard computational hardware to run the fiducial software,
and running the fiducial software on the tablet is difficult and gives no computational insight.
Therefore, we offload single video frames from the tablet to a companion board for analysis,
and we use a Raspberry Pi 4 as it is representative of the hardware that we will eventually embed
into a drone for onboard processing.
As the tablet connects to the DJI controller using its only wired interface,
we offload the video frames to the companion board via TCP ports
over an ad-hoc WiFi network hosted by the companion board,
after compressing the frames into \texttt{.webp} format in grayscale at 20\% quality.
This resulted in a maximum framerate of about 7 Hz with a delay of between 0.5 and 2 seconds
from image acquisition to command delivery.
The drone executes the following mission autonomously:
\begin{enumerate}
	\item takeoff to an altitude of 2 meters
	\item search for the landing pad by spinning in place and tilting the gimbal up and down
	\item approach the landing pad once acquired, track it with gimbal tilt and the drone yaw
	\item align to the yaw of the landing pad once with a small distance from it
	\item descend until touchdown; disable the motors
\end{enumerate}

The system is able to accomplish its original goal of tracking the marker over time
by actuating the gimbal,
as shown in Fig.~\ref{figure:tracking_example}.
The $x,y$ pixel centers of the landing pad $u_n, v_n$ respectively are normalized to $[-1, 1]$
and remain close to 0 throughout the example landing, meaning that the landing pad stays near the
center of the video frames.

Four of the five systems show 20 successful landings;
their
accuracy shown in Fig.~\ref{figure:landing_radii}.
Two factors influence the accuracy: the prevalence of orientation ambiguity and the ability of the drone
to maintain its pose estimate throughout the entire landing.
April Tag 48h12 and 24h10 both allow the drone to detect the landing pad
until near touchdown,
but their significantly different rates of orientation ambiguity
result in very
different landing accuracies.
The WhyCode systems provide similar accuracies, but with both being less accurate
than April Tag 48h12 because of
the relatively high altitude ($0.6$ meters) at which the the landing pad becomes too large for
the camera frame, forcing the drone to commit earlier to finish the landing blind.
WhyCode Multi achieved no successful landings,
but overshot the landing pad.

\begin{figurehere}
	\centering
	\includegraphics[width=\linewidth]{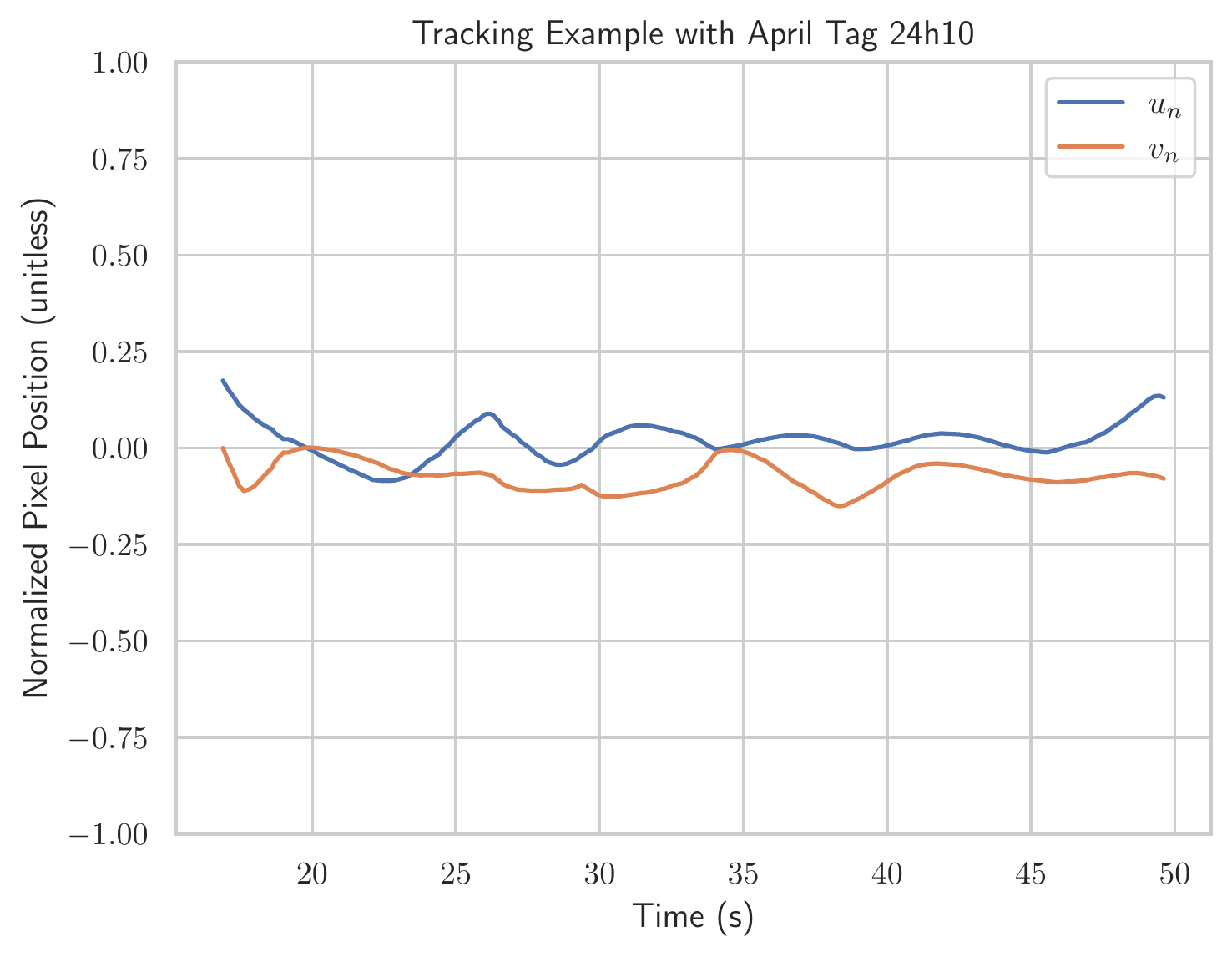}
	\caption{The normalized pixel positions of the landing pad during an example landing.
		 The drone is able to track the marker with yaw and gimbal tilt,
		 keeping $(u_n, v_n) \approx (0, 0)$.}
	\label{figure:tracking_example}
\end{figurehere}
\vspace{-0.35cm}
\begin{figurehere}
	\centering
	\includegraphics[width=\linewidth]{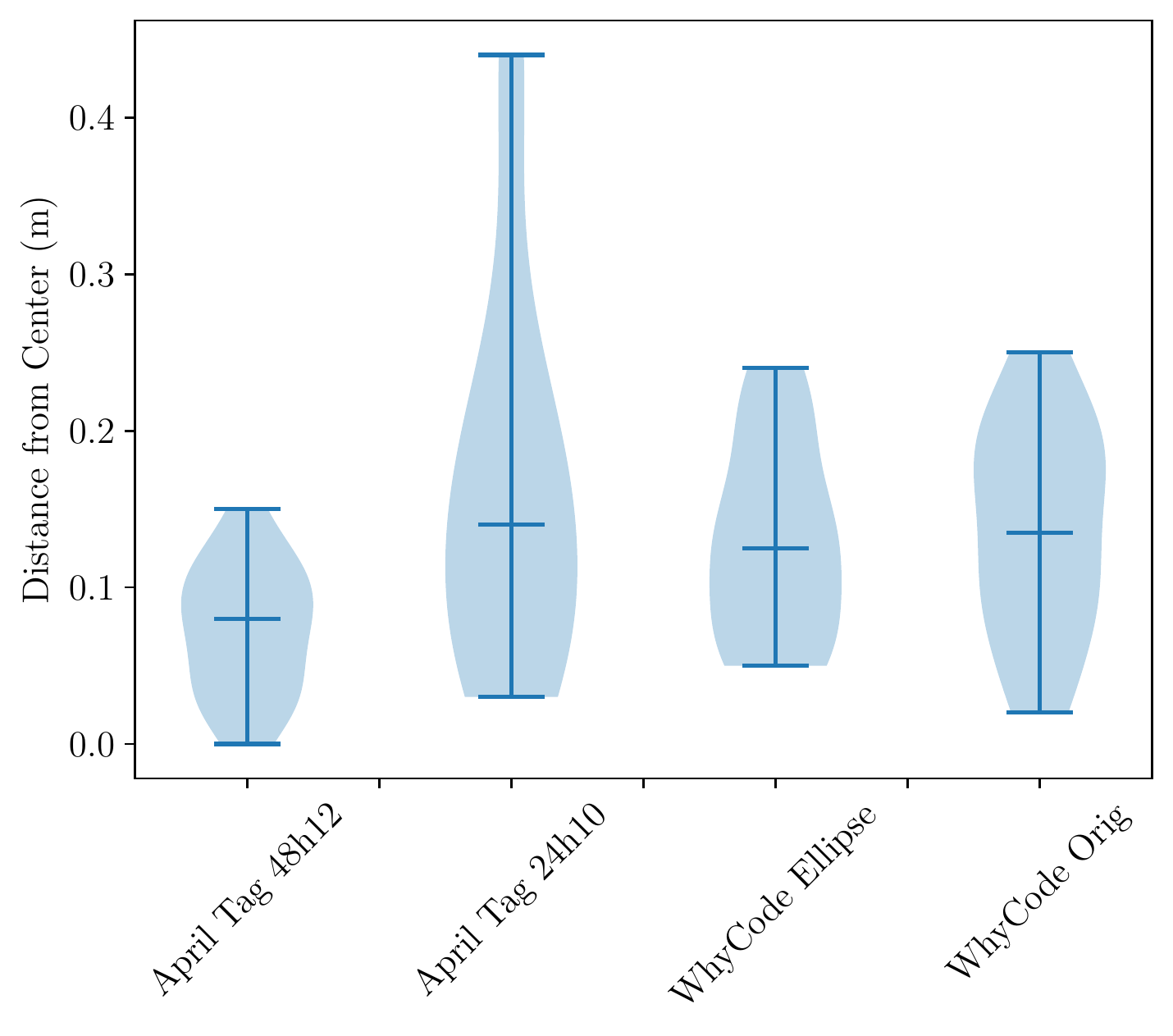}
	\caption{Distances from the camera to the center of the landing pad after touchdown for each successful system.}
	\label{figure:landing_radii}
\end{figurehere}
\vspace{-0.35cm}
\noindent

Finally, Fig.~\ref{figure:control_example_discontinuities} shows an example of erroneous
commands generated by the system as a result of the markers' orientation ambiguity.
The pitch and roll components of the VirtualStick commands can exhibit the same discontinuities
as the position targets, resulting in sign flips where the landing pad appears to jump from one
side of the drone to the other.
In this case, the drone oscillates to the left and right of the landing pad 4 times before
converging to an accurate enough position for descent.
This demonstrates that the orientation ambiguity problem does not prohibit the drone from landing
even with an actuated camera.

\begin{figurehere}
	\centering
	\includegraphics[width=\linewidth]{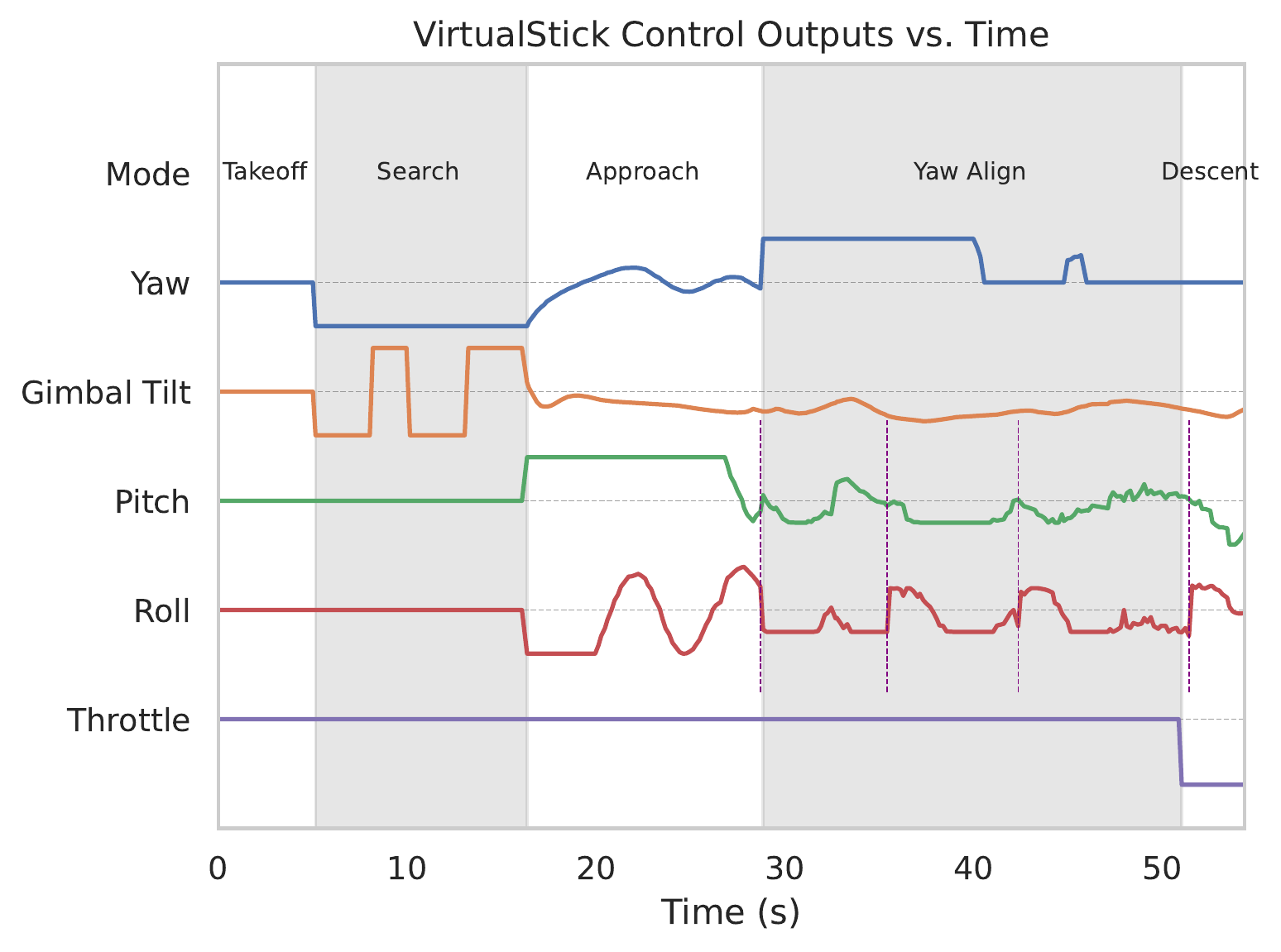}
	\caption{Example of erroneous control signals (marked by vertical lines) generated by orientation ambiguity.}
	\label{figure:control_example_discontinuities}
\end{figurehere}
\vspace{-0.35cm}

\subsection{Future Work}

We will test two further facets of the fiducial landing method
-- both in an effort to reduce the effects of orientation ambiguity.
The first change is to use essentially the same setup with the addition of a typical Kalman filter
to smooth the position targets.
We anticipate that this will only partially mitigate the effects of orientation ambiguity,
but still has the potential to increase the landing accuracy while only using data from the fiducial
systems.
The second change is to use only the position of the landing pad directly from the fiducial system,
and to derive its pitch and roll from the orientation of the camera under the assumption that the
landing pad is flat on the ground.
The orientation of the camera is available from the DJI SDKs, but does not generalize to all systems,
as some gimbals and cameras do not offer this data natively.
We anticipate that this method will remove the effects of orientation ambiguity entirely.
We will conduct these subsequent experiments outdoors on a larger drone platform
at longer distances.


%% file: sections/lava_flow_landing.tex
\label{section:lava_landing}

In collaboration with the \gls{RAVEN} project~\cite{raven_whitepaper},
we are developing a method for identifying viable landing sites in lava environments using only
onboard, embedded processing.
The goal of \gls{RAVEN} is to use Iceland as an analog environment for testing the research gains
made by using collaborative drone-rover teams in the context of Mars
exploration.
One of the possible benefits of using a drone in an exploration team is the ability to quickly
reach and collect data at sites that are difficult for the rover to reach
-- such as on top of a solidified lava flow.
Analog missions are essentially a secondary form of simulation,
where everything is more realistic than in a computer simulation,
but not perfectly similar to actual mission conditions.
Thus, in the field work for \gls{RAVEN}
-- which takes place near a lava flow called Holuhraun in the Icelandic highlands --
the missions can be manually executed.
This is particularly true for the lava landings, as the waypoint-to-waypoint flight
and collection of pictures can be easily automated.
The \gls{RAVEN} fieldwork
is a \emph{relatively} inexpensive way to test autonomous landing
in a convincing analog environment that gives good insight
into the challenges of accomplishing lava landing on Mars.

\subsection{Landing Site Identification}

This scenario requires a more customized method for identifying safe landing sites than the
fiducial methods, with their well-tested systems.
However, many of the requirements are the same:
the system must identify a \emph{safe} landing area,
in a stable way (i.e.~no bouncing or discontinuities as in Section~\ref{section:fiducial_landing}),
and must be able to safely search for a landing target (i.e.~by actuating the camera instead of only
translating through its environment).
We use a RealSense D455 \gls{RGBD} camera,
which produces a synced stream of
RGB and depth images, accelerometer data, and gyroscope data.
Our drone platform (Fig.~\ref{figure:depth_drone_flying_stora_bolluhraun})
is a Tarot 680 hexacopter with a Cube Orange flight controller,
a HereLink telemetry system, and many custom, 3D-printed parts for shielding electronic components
and batteries from harsh Icelandic weather and for modifying the landing gear.
The D455 is mounted on a custom gimbal to allow the drone to properly stabilize and actuate it.
It has a Raspberry Pi 4 with 2 GB of RAM as a companion board,
and the possiblity of adding edge computing boards:
a Google Coral USB TPU or an NVIDIA Jetson Nano.
Our main analog environment, Holuhraun (meaning ``Hole Lava'', shown in Fig.~\ref{figure:holuhraun}),
is reachable once a year,
so we test in lava fields surrounding the nearby town of Hafnarfjör{\dh}ur.
(Fig.~\ref{figure:depth_drone_flying_stora_bolluhraun}).

\vspace{0.1cm}
\begin{figurehere}
	\centering
	\includegraphics[width=0.9\linewidth]{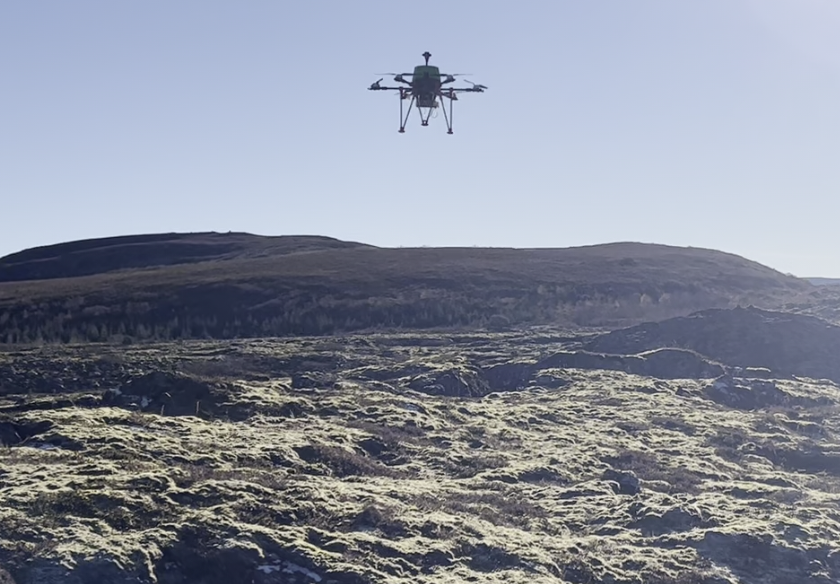}
	\caption{The Tarot 680 drone with RealSense D455.}
	\label{figure:depth_drone_flying_stora_bolluhraun}
\end{figurehere}
\vspace{-0.35cm}
\noindent
The most favorable landing sites in lava flows are large, relatively flat and smooth
areas called lava-rise plateaus (or inflation plateaus)~\cite{holuhraun_characterization}.
The main computational task is to analyze depth images to avoid rough areas such as
in Fig.~\ref{figure:holuhraun} left,
and to identify smooth areas such in Fig.~\ref{figure:holuhraun} right,
while avoiding the cracks.
After the landing target is identified,
the approach can be similar to that of Section~\ref{section:fiducial_landing}.

\vspace{0.1cm}
\begin{figurehere}
	\centering
	\includegraphics[width=0.475\linewidth]{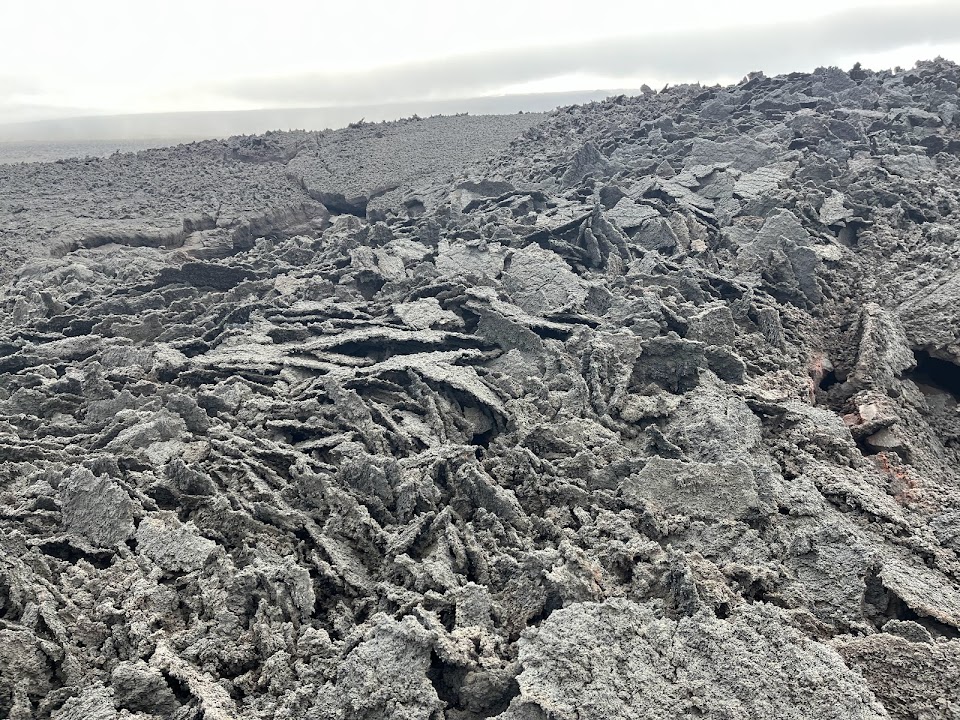}
	\includegraphics[width=0.475\linewidth]{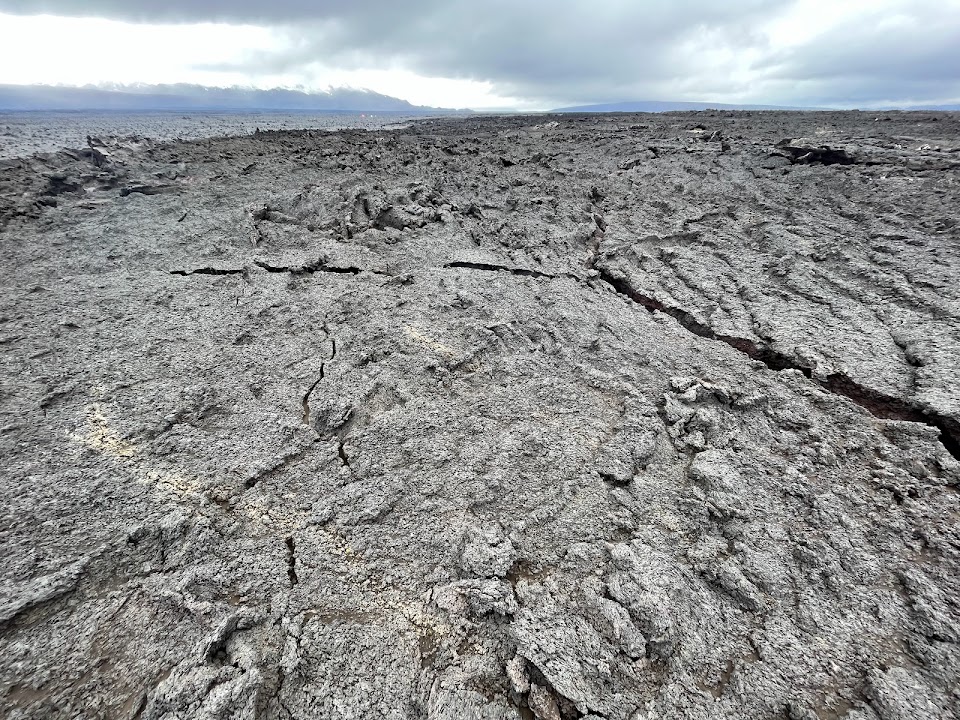}
	\caption{Holuhraun's varied terrains.}
	\label{figure:holuhraun}
\end{figurehere}
\vspace{-0.35cm}

In both Holuhraun and the Hafnarfjör{\dh}ur lava fields, we have collected data from the D455
for offline testing of our experimental terrain analysis methods.
We have collected data from the D455 that includes the streams of all of its sensors,
so that we can develop a solution for terrain analysis offline.
Lava flows in Iceland tend to have very little obstructive vegetation or dynamic obstacles,
so we can focus on efficient, static terrain analysis.

\subsection{Solution Development \& Testing}

The first step is to correct for the orientation of the D455
which is not always pointed straight down.
We calculate the orientation of the D455 from its accelerometer and gyroscope data,
and transform the RGBD images by rotating them by the inverse of its pitch and roll.
We ignore the yaw because the sensor's yaw is aligned with that of the drone.

The next step is to process the properly oriented RGB and depth images
to identify adequately large, smooth, level regions
and to avoid rough areas, cracks and pits.
So far we develop in Python 3 and depend on the RealSense libraries, OpenCV, and Numpy.
Subsequent implementations will be in C++ with a focus on efficiency and real time execution.

We have experimented with feature detection algorithms on both the RGB image and the depth image,
e.g.~applying scale-invariant feature transforms (SIFT)\cite{lowe99:_objec}
to both images to find common features.
Since the RGB images have many specks of light and dark,
SIFT identifies many key points (Fig.~\ref{figure:sift_detection} left, each circle represents a key point).
Conversely, the corresponding depth images are smooth, and SIFT detects fewer key points (Fig.~\ref{figure:sift_detection} right).
There seldom are common key points.
Other feature extraction algorithms fare similarly, encouraging us to develop our own method.
This solution runs at $\geq1.7$ Hz with default parameters on our Raspberry Pi.

\vspace{0.1cm}
\begin{figurehere}
	\centering
	\includegraphics[width=0.45\linewidth]{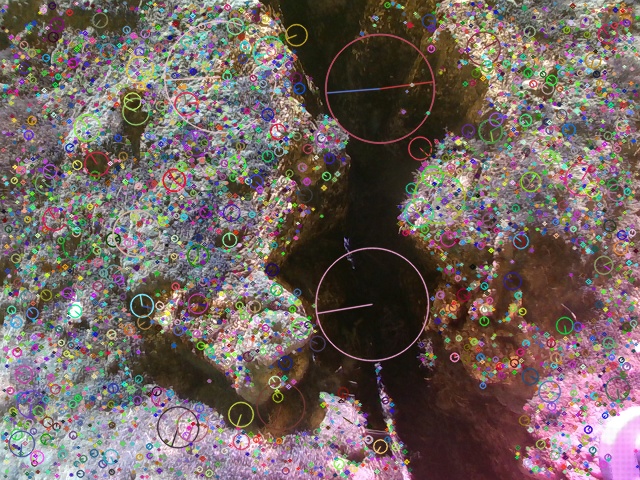}
	\includegraphics[width=0.45\linewidth]{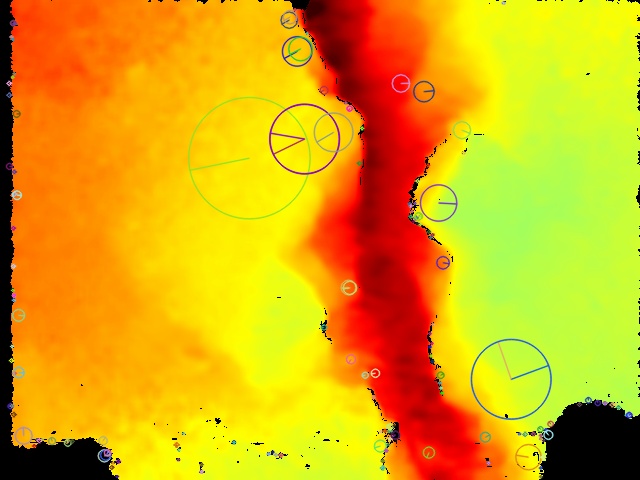}
	\caption{Example SIFT detection.
		 Left: RGB.
		 Right: depth (post-processed).}
	\label{figure:sift_detection}
\end{figurehere}
\vspace{-0.35cm}

Initial experiments (Fig.~\ref{figure:crack_detection})
show some success in crack detection by eliminating areas of high slope by applying a Gaussian filter
and a threshold on the absolute value of the terrain slope, (approximated via 2D discrete difference).
This solution runs at $\geq8$ Hz on our the Raspberry Pi.

\vspace{0.1cm}
\begin{figurehere}
	\centering
	\includegraphics[width=0.90\linewidth]{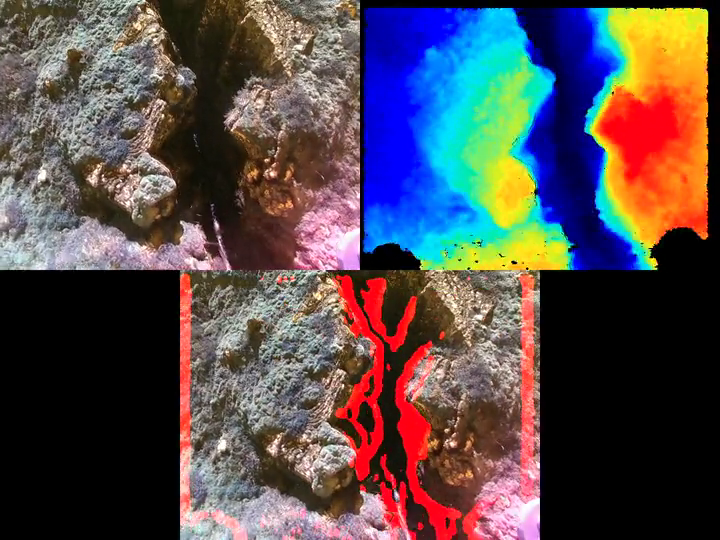}
	\caption{Example crack detection.
		 Top left: RGB.
		 Top right: depth (post-processed).
		 Bottom: RGB with hazardous areas labeled in red.}
	\label{figure:crack_detection}
\end{figurehere}
\vspace{-0.35cm}

Robust plane fits, as mentioned by Scherer et.~al~\cite{fullscale_helicopter_robust_plane_fit}
are a promising method, as our landing scenario does not involve dynamic or
compliant obstacles.
Deep learning methods -- particularly \gls{CNN}s or U-nets --
are also promising methods that could combine information from both the
RGB and depth images, and our edge computing boards are designed to run such algorithms
efficiently.
For this, we would import a \gls{DTM} of Holuhraun (from the
2022 RAVEN fieldwork), into AirSim~\cite{10.1007/978-3-319-67361-5_40}.
Then, we would label regions of the \gls{DTM} as suitable or unsuitable for landing
and generate a synthetic training data set of labeled \gls{RGBD} images.

The last step is to track the landing site, by actuating the camera (if beneficial),
and translating the location of the landing site into a target velocity or position
so that the drone can approach.

We will test any successful terrain analysis methods on physical hardware in order to determine its
power consumption and runtime framerate,
then deploy the system near Hafnarfjör{\dh}ur for real world landing tests,
with a final test at Holuhraun.

%% file: sections/conclusions_and_future_work.tex
Our goal is to improve autonomous drone landing methods
based on fiducial systems
and to develop a new method for landing in solidified lava fields.
We have made advancements on the fiducial front by actuating the drone's monocular camera to track
marked landing pads independently of the drone's orientation,
increasing the robustness of the landing pad detection.
However, this complicates the estimation of the drone's pose relative to the
landing pad because of marker orientation ambiguity.
Thus, we evaluated the prevalence of
orientation ambiguity in multiple fiducial systems
and some techniques to mitigate it,
and we will continue by filtering the pose estimates
and deriving the marker orientation using the camera's \gls{IMU}.
%
For lava flow landing,
we have built a custom drone platform that uses an \gls{RGBD}
camera to analyze the terrain, 
and used it to collect data in Icelandic lava fields
to develop our terrain analysis methods offline.
These methods have shown success in 
identifying phenomena
such as cracks as unsafe landing sites,
but we will test more such techniques for identifying more terrain features,
focusing on efficiency
and real-time execution.
Finally, we will test successful methods in real world autonomous landing scenarios.